\documentclass[10pt,twocolumn,letterpaper]{article}

\usepackage{iccv}              

\usepackage{algorithm}
\usepackage{algorithmicx}
\usepackage{algpseudocode}
\makeatletter
\algnewcommand{\INPUT}[1]{%
  \State \textbf{Input:} #1
}
\algnewcommand{\OUTPUT}[1]{%
  \State \textbf{Output:} #1
}
\makeatother

\usepackage{microtype}
\usepackage{graphicx}
\usepackage{subcaption}
\usepackage{booktabs}
\usepackage{amsmath}
\usepackage{amssymb}
\usepackage{mathtools}
\usepackage{amsthm}
\usepackage[T1]{fontenc}
\usepackage[utf8]{inputenc}

\usepackage{listings}
\usepackage{xcolor}
\usepackage{subcaption}
\usepackage[normalem]{ulem}

\usepackage{soul}

\usepackage{xcolor}  
\usepackage{colortbl}  

\definecolor{keyword}{rgb}{0.55, 0.0, 0.0}  
\definecolor{comment}{rgb}{0.25, 0.5, 0.25}  
\definecolor{string}{rgb}{0.58, 0, 0.82}  
\definecolor{background}{rgb}{0.95, 0.95, 0.95}  
\definecolor{lightred}{rgb}{1.0, 0.6, 0.6}  
\definecolor{lightgreen}{rgb}{0.6, 1.0, 0.6}  

\lstdefinestyle{pythonstyle}{
  language=Python,
  commentstyle=\color{comment}\itshape,
  keywordstyle=\color{keyword}\bfseries,
  stringstyle=\color{string},
  basicstyle=\ttfamily\small,
  numberstyle=\tiny\color{gray},
  numbers=left,
  numbersep=5pt,
  showstringspaces=false,
  showspaces=false,
  showtabs=false,
  breaklines=true,
  breakatwhitespace=true,
  moredelim=**[is][\color{red}]{-}{-},  
  moredelim=**[is][\color{green}]{+}{+},   
  tabsize=2,
  frame=single,
  captionpos=b,
  xleftmargin=20pt,
  xrightmargin=20pt,
  framexleftmargin=10pt,
  framexrightmargin=10pt
}

\usepackage[textsize=tiny]{todonotes}

\usepackage{multirow}
\usepackage{cuted}
\usepackage{ulem}  

\definecolor{iccvblue}{rgb}{0.21,0.49,0.74}
\usepackage[pagebackref,breaklinks,colorlinks,allcolors=iccvblue]{hyperref}

\def\paperID{*****} 
\def\confName{ICCV}
\def\confYear{2025}

\title{Training-free and Adaptive Sparse Attention for Efficient Long Video Generation}

\author{
\textbf{Yifei Xia}$^{1,2}$ \quad
\textbf{Suhan Ling}$^{1,2}$ \quad
\textbf{Fangcheng Fu}$^{1}$ \quad
\textbf{Yujie Wang}$^{1,2}$ \quad
\\
\textbf{Huixia Li}$^{2}$ \quad
\textbf{Xuefeng Xiao}$^{2}$ \quad
\textbf{Bin Cui}$^{1}$ \quad
\\
$^{1}$Peking University \quad $^{2}$ByteDance
\\
\{yifeixia, lingsuhan\}@stu.pku.edu.cn \quad
\{ccchengff, alfredwang, bin.cui\}@pku.edu.cn \\
\{lihuixia, xiaoxuefeng.ailab\}@bytedance.com
}

\begin{document}

\maketitle

\begin{strip}
\centering
\includegraphics[width=\textwidth]{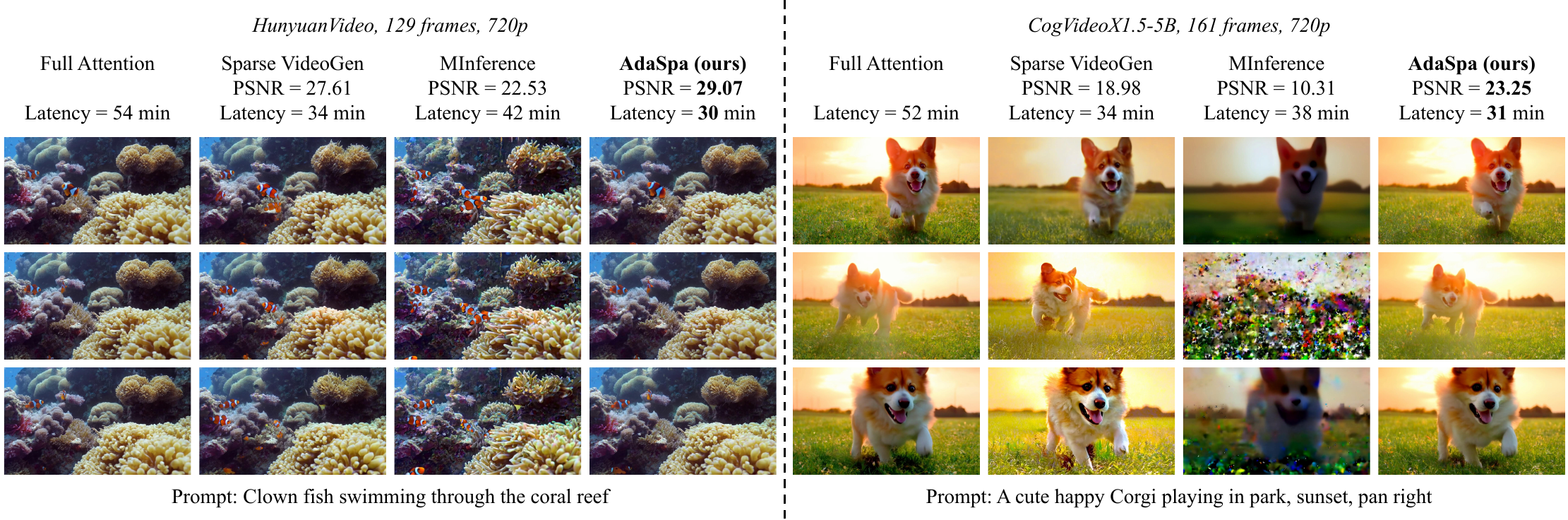}
\captionof{figure}{Comparison of the visualization effects of different sparse attention methods on \emph{HunyuanVideo}~\cite{kong2024hunyuanvideo} and \emph{CogVideoX1.5-5B}~\cite{yang2024cogvideox}. Our method \emph{AdaSpa} consistently achieves the best performance and the best speedup, and keep almost the same as original videos.}
\label{fig:Visual}
\end{strip}
\begin{abstract}
Generating high-fidelity long videos with Diffusion Transformers (DiTs) is often hindered by significant latency, primarily due to the computational demands of attention mechanisms. 
For instance, generating an 8-second 720p video (110K tokens) with HunyuanVideo takes about 600 PFLOPs, with around 500 PFLOPs consumed by attention computations. 
To address this issue, we propose \textbf{AdaSpa}, the first \textbf{Dynamic Pattern} and \textbf{Online Precise Search} sparse attention method.
Firstly, to realize the Dynamic Pattern, we introduce a blockified pattern to efficiently capture the hierarchical sparsity inherent in DiTs. 
This is based on our observation that sparse characteristics of DiTs exhibit hierarchical and blockified structures between and within different modalities.
This blockified approach significantly reduces the complexity of attention computation while maintaining high fidelity in the generated videos.
Secondly, to enable Online Precise Search, we propose the Fused LSE-Cached Search with Head-adaptive Hierarchical Block Sparse Attention. 
This method is motivated by our finding that DiTs’ sparse pattern and LSE vary w.r.t. inputs, layers, and heads, but remain invariant across denoising steps.
By leveraging this invariance across denoising steps, it adapts to the dynamic nature of DiTs and allows for precise, real-time identification of sparse indices with minimal overhead.
AdaSpa is implemented as an \textbf{adaptive, plug-and-play solution} and can be integrated seamlessly with existing DiTs, requiring neither additional fine-tuning nor a dataset-dependent profiling. Extensive experiments validate that AdaSpa delivers substantial acceleration across various models while preserving video quality, establishing itself as a robust and scalable approach to efficient video generation.
\end{abstract}    
\section{Introduction}
\label{sec:intro}
Diffusion models~\cite{ddpm,ddim,diffusionbeatgan,cfg,stablediffusion} have emerged as a powerful framework for generative tasks, achieving state-of-the-art results across diverse modalities, including text-to-image synthesis~\cite{sauer2024fast,flux2024,chen2024pixartdelta,zhang2023adding,schwartz2023discriminative,cao2024leftrefill,fei2024dysen,qu2024discriminative,shirakawa2024noisecollage,xue2024accelerating,li2024distrifusion}, realistic video generation~\cite{genmo2024mochi,kong2024hunyuanvideo,yang2024cogvideox,hong2022cogvideo,lin2024open,zheng2024open,jiang2024videobooth,wu2023tune,ho2022video}, and 3D content creation~\cite{chen2024vp3d,chen2024sculpt3d,poole2022dreamfusion,huang2024dreamcontrol,hollein2024viewdiff}. Recently, the introduction of Diffusion Transformers (DiTs)~\cite{peebles2023scalable}, exemplified by Sora~\cite{videoworldsimulators2024}, has set new benchmarks in video generation, enabling the production of long, high-fidelity videos. 

\begin{figure}[!t]
    \centering
    \includegraphics[width=\linewidth]{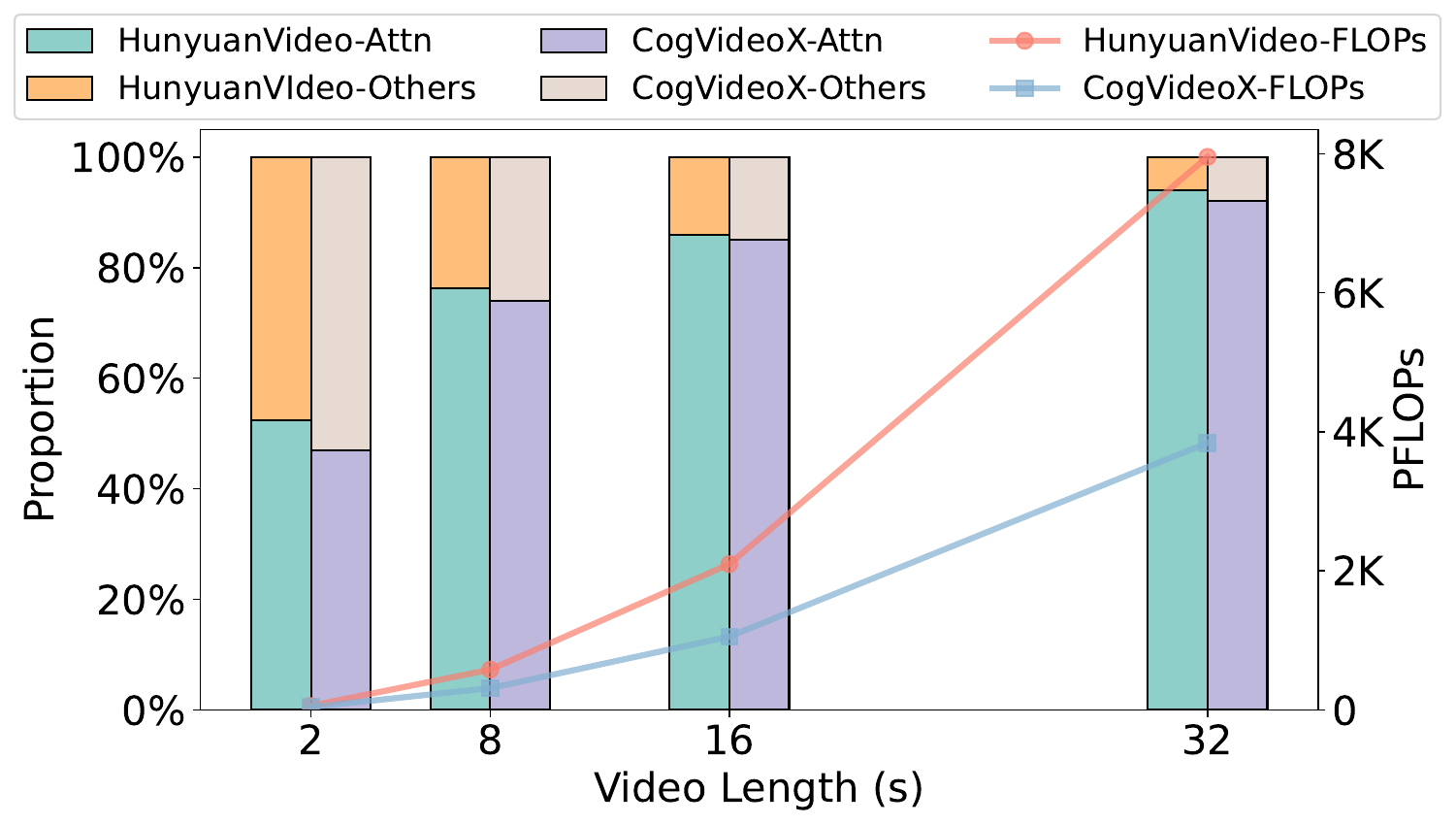}
    \caption{The total FLOPs required and the proportion of attention when generating 720p videos with different video lengths (16FPS).
    }
    \label{fig:attn_proportion}
\end{figure}

Despite these advances, generating high-quality videos remains computationally expensive, especially for long videos~\cite{chen2025ouroboros,tan2024video,fang2024pipefusion,he2022latent}. The attention mechanism~\cite{vaswani2017attention} in the Transformer architecture~\cite{vaswani2017attention}, with its $O(n^2)$ complexity, is a major bottleneck, where $n$ denotes the sequence length. 
For instance, generating an 8-second 720p video with HunyuanVideo takes about 600 PFLOPs, with nearly 500 PFLOPs consumed by attention computations.
This proportion increases with higher resolution or longer duration videos, as illustrated in Figure~\ref{fig:attn_proportion}.

\begin{figure*}[!t]
    \centering
    \includegraphics[width=1.0\linewidth]{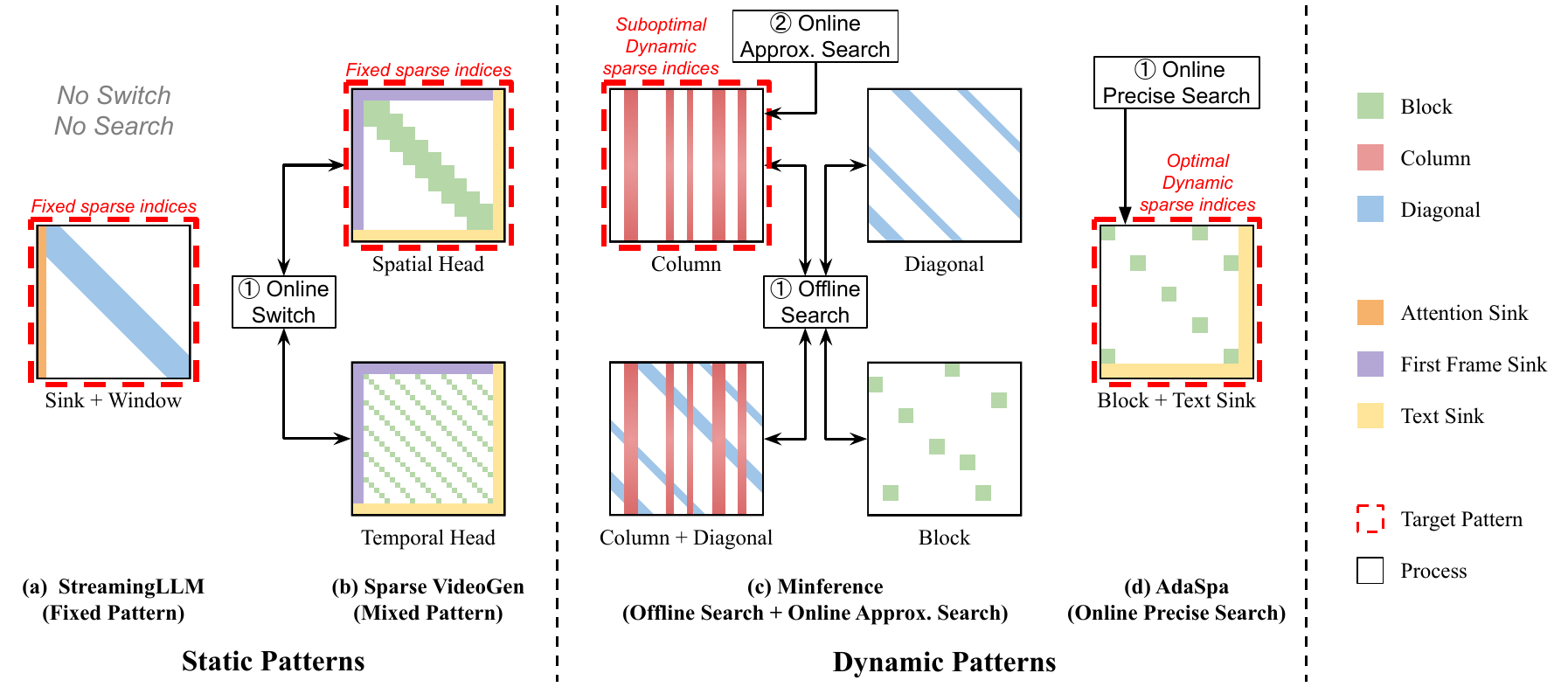}
    \caption{Different types of Sparse Pattern recognition methods. (a) StreamingLLM: using a static \emph{sink}+\emph{sliding window} pattern, need no search or switch. (b) Sparse VideoGen: preparing two predefined Static Patterns, and using an online switching method to determine which to use. (c) MInference: preparing several dynamic patterns, first do an offline search to determine the target pattern to use, then perform an online approximate search to search suboptimal sparse indices of this pattern. (d) AdaSpa: our method proves that the most suitable pattern for DiT is \emph{blockified} pattern, and performs an online precise search to find the optimal sparse indices for blockified pattern. 
    }
    \label{fig:patterns}
\end{figure*}

Although attention mechanisms are essential for sound performance, they involve significant computational redundancy~\cite{child2019generating}. Addressing this redundancy can greatly reduce inference costs and accelerate video generation~\cite{xi2025sparse}. Sparse attention mechanisms~\cite{ding2025efficient,zhang2025fast,tan2025dsv,yuan2025native,fu2024moa,lu2025moba,gao2024seerattention,pu2025efficient,wang2024qihoo,jiang2024minference,liu2021transformer,acharya2024star,han2023lm,zaheer2020big,beltagy2020longformer,xi2025sparse,qiu2019blockwise,child2019generating}, which exploit this redundancy, have shown success in large language models (LLMs) by reducing computational costs without compromising performance. 
Sparse attention typically characterizes this redundancy as \emph{sparse patterns} (a.k.a. \emph{sparse masks}), indicating which interactions between tokens can be omitted to reduce computational load. 
The specific positions of the selected tokens in \emph{sparse patterns} that are not omitted are called \emph{sparse indices}.
Based on the flexibility of pattern recognition, existing \emph{sparse patterns} can be broadly categorized into the following two types: 
\begin{itemize}
    \item \textbf{Static Pattern}~\cite{acharya2024star,han2023lm,zaheer2020big,beltagy2020longformer,xi2025sparse,child2019generating} refers to the use of predetermined sparse indices that are defined by prior knowledge. 
    This category can be further divided into two types:
    
    \underline{\textit{Fixed Pattern}} uses only one fixed sparse pattern based on empirical experience. For instance, LM-Infinite~\cite{han2023lm} and StreamingLLM~\cite{xiao2309efficient} (Figure~\ref{fig:patterns}a) consistently utilize the sliding window~\cite{beltagy2020longformer} pattern. This approach is straightforward, generally requiring no pattern search, and only necessitates the prior specification of hyperparameters.
    
    \underline{\textit{Mixed Pattern}} involves determining several fixed patterns based on experience and then selecting one or more of these patterns during the execution of attention. Examples include BigBird~\cite{zaheer2020big} and Sparse VideoGen~\cite{xi2025sparse} (Figure~\ref{fig:patterns}b), which typically perform a rough online switching mechanism to estimate and determine which pattern (or combination of patterns) should be applied in each attention operation. 
    
    \item \textbf{Dynamic Pattern}~\cite{tan2025dsv,gao2024seerattention,pu2025efficient,jiang2024minference,liu2021transformer,qiu2019blockwise} features ad hoc sparse indices that need to be decided in real time. Examples include DSA~\cite{liu2021transformer} and MInference~\cite{jiang2024minference} (Figure~\ref{fig:patterns}c). 
    It necessitates a search to determine which indices to use for each attention operation.
    Due to the extensive time consumption involved in searching, current Dynamic Pattern methods typically rely on offline search and/or online approximation search.
    
    \underline{\textit{Offline Search}} methods involve performing offline searches to determine the specific indices. A subset of the target dataset is usually used in the offline search.
    
    \underline{\textit{Online Approximate Search}} methods involve searching in real-time, yet applying some form of approximation to estimate \emph{sparse indices} during the execution.

    
\end{itemize}

However, due to the dynamic complexity and data-adaptive nature of DiT patterns, these methods face significant limitations 
when applied to DiTs. 

\textit{Firstly, the \textbf{Static Pattern} is not flexible enough to summarize the sparse characteristics of DiTs.} 
In particular, as we will show in Section~\ref{sec:observation}, the sparse patterns of DiTs are extremely dynamic and irregular. Thus, static pattern methods fail to accurately capture the sparse indices and thereby suffer from poor performance (as evaluated in Section~\ref{sec:experiments}).
%

\textit{Secondly, the existing \textbf{Dynamic Pattern} methods are unable to adaptively and accurately identify the sparse patterns of DiTs.}
For one thing, our empirical observations in Section~\ref{sec:observation} demonstrate that the sparsity of DiTs exhibits considerable variation depending on the input, which makes offline search in DiTs lack good portability and accuracy.
For another, it can be observed that the sparse indices in DiTs are complex, with key areas being dispersed and not concentrated and continuous, making it difficult to accurately estimate sparse indices through approximation search. 
Thus, directly applying current dynamic pattern methods (e.g., MInference) to DiT also yields poor results (detailed in Section~\ref{sec:experiments}).

Therefore, identifying and generalizing \emph{sparse patterns} suitable for DiTs, and implementing kernel-efficient methods for precise pattern search and attention execution remains an urgent problem to be solved.


Motivated by this, we propose \textbf{\underline{Ada}ptive \underline{Spa}rse Attention (AdaSpa)}, the first Dynamic Pattern + Online Precise Search (Figure~\ref{fig:patterns}d) method for high-fidelity sparse attention. It is a training-free and data-free method designed to accelerate video generation in DiTs while preserving generation quality. It outperforms all other SOTA methods in both Static and Dynamic Patterns, as shown in Figure~\ref{fig:Visual}. Our contributions are summarized as follows:

\begin{itemize}
    \item \textbf{Comprehensive Analysis of Attention Sparsity in DiTs.} We present an in-depth analysis of sparse characteristics in attention mechanisms for DiTs, examining the special sparse characteristics of DiTs to reveal optimal sparsity strategies and provide new insights for future research. 
    Based on extensive observations and summaries, we found that the sparse characteristics of DiTs have two traits: 1) \textbf{Hierarchical} and \textbf{Blockified}, 2) \textbf{Invariant in steps, Adaptive in prompts and heads}.
    
    \item \textbf{First Dynamic Patterns and Precise Online Search Sparse Attention Solution without Training and Profiling.} We propose AdaSpa, a novel sparse attention acceleration framework that is both training-free and data-free. As shown in Figure~\ref{fig:patterns}d, AdaSpa is the first effective method that combines Dynamic Pattern and Online Precise Search, proposing an efficient pipeline for online sparse pattern search and fine-grained sparse attention computation. Leveraging the invariant characteristics across denoising steps, AdaSpa is equipped with Fused LSE-Cached Online Search, which reduces online search time to under 5\% of full attention generation time using our optimized kernel, significantly reducing the additional time for search while ensuring accurate search. Additionally, in order to better adapt to the sparse characteristics of DiT, we propose a Head-Adaptive Hierarchical Block Sparse method for AdaSpa to address the head-adaptive sparsity feature of DiTs.
    
    \item 
    \textbf{Implementation and Evaluation.}
    AdaSpa provides a plug-and-play $adaspa\_attention\_handler$ that seamlessly integrates with DiTs, requiring no fine-tuning or data profiling. It is orthogonal to other acceleration techniques like parallelization, quantization and cache reuse. Extensive experiments validate AdaSpa's consistent speedups across models with negligible quality loss. 
\end{itemize}

\section{Preliminaries}
\label{sec:pre}

\subsection{Diffusion Transformers and 3D Full Attention}







Diffusion Transformers (DiTs)~\cite{peebles2023scalable} refine predictions with a diffusion process, handling multimodal data like video and text through an attention mechanism that captures spatial, temporal, and cross-modal dependencies.
\begin{figure}[!t]
    \centering
    \includegraphics[width=\linewidth]{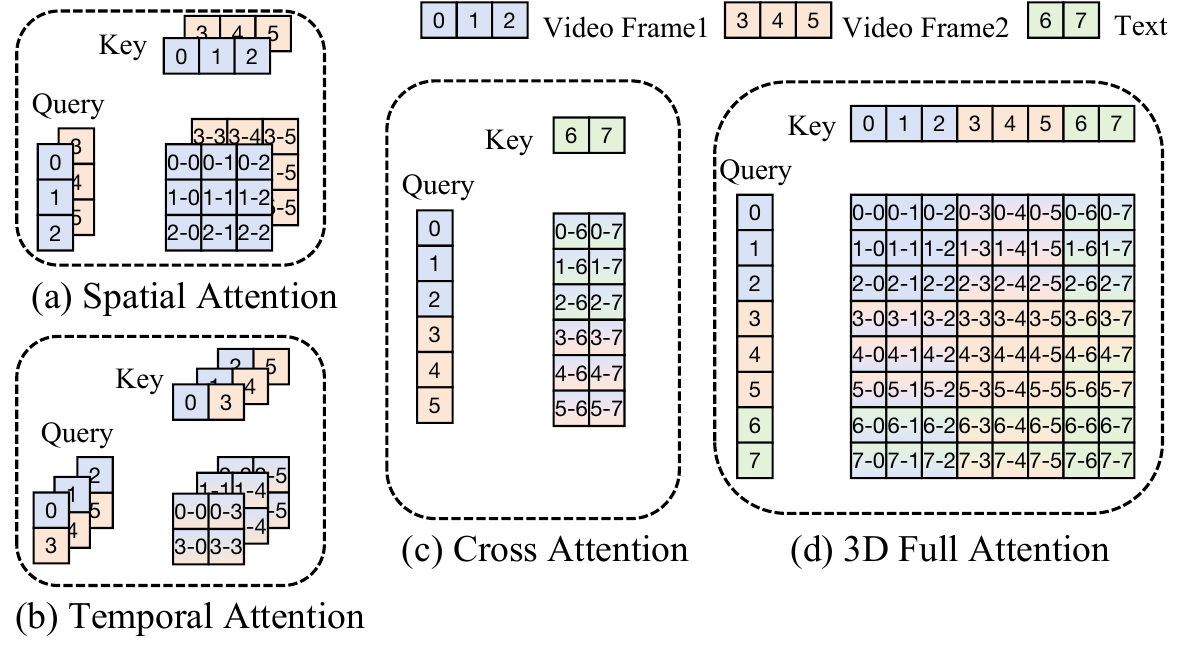}
    \caption{Different Attention Mechanisms in DiTs.}
    \label{fig:attention}
\end{figure}
DiTs traditionally use Spatial-Temporal Attention~\cite{zheng2024open,lin2024open}, applying spatial attention with each video frame, temporal attention across all frames, and cross-attention to connect video and text, as shown in Figure~\ref{fig:attention}. This separation limits frame continuity and fusion.

Figure~\ref{fig:attention}d illustrates the 3D Full Attention mechanism~\cite{hong2022cogvideo,yang2024cogvideox,kong2024hunyuanvideo} in DiTs. It integrates video and text tokens into a unified sequence and applies self-attention across them. Operating in the latent space, DiTs process video frames that have been pre-encoded.
Let \( f \) be the number of latent frames, \( h \times w \) the spatial resolution of each frame, and \( t \) the text token length, with \( f \cdot h \cdot w \gg t \). The total sequence length, \( L \), can be represented as:
\begin{equation}
\label{eq:seqlen}
L = f \cdot h \cdot w + t.
\end{equation}
This unified approach enhances modality fusion and boosts overall performance.

Despite the increased computational cost of 3D Full Attention, it marks the future of DiTs, offering superior multimodal learning compared to Spatial-Temporal Attention.

\subsection{FlashAttention}
\label{subsec:flash}


In the self-attention mechanism \cite{vaswani2017attention}, tokens are projected into the query, key, and value matrices \( \mathbf{Q}, \mathbf{K}, \mathbf{V} \in \mathbb{R}^{H \times L \times D} \), where \( H \) is the number of attention heads, \( L \) is the input length, and \( D \) is the hidden dimension of each head. The attention weights matrix \( \mathbf{W}_{\text{attn}} \in \mathbb{R}^{L \times L} \) is computed as:
\begin{equation}
\mathbf{W}_{\text{attn}} = \text{softmax}\left(\frac{\mathbf{Q} \mathbf{K}^\top}{\sqrt{D}}\right),
\end{equation}
which quantifies token-to-token interactions across the sequence. 
To maintain numerical stability during the exponentiation, 
the Log-Sum-Exp (LSE)~\cite{bahdanau2016neuralmachinetranslationjointly} trick is commonly employed. 
Let $\mathbf{Z} = \tfrac{\mathbf{Q}\mathbf{K}^\top}{\sqrt{d}}$ and denote by $z_j$ the $j$-th component of a row $\mathbf{z}$. 
Then, LSE can be written as:
\begin{equation}
\begin{aligned}
\mathrm{LSE}(\mathbf{z}) 
&= \log \sum_{j} \exp(z_j) \\
&= \max_j z_j + \log \sum_{j} \exp\bigl(z_j - \max_k z_k\bigr).
\end{aligned}
\end{equation}
Using this, the safe Softmax can be expressed as:
\begin{equation}
\mathrm{Softmax}_{\mathrm{safe}}(z_j)
\;=\; \exp\bigl(z_j \;-\;\mathrm{LSE}(\mathbf{z})\bigr),
\end{equation}
and the entire dense attention distribution in a numerically stable form is:
\begin{equation}
\mathbf{W}_{\text{attn}}
\;=\; \text{Softmax}_{\mathrm{safe}}(\frac{\mathbf{Q}\mathbf{K}^\top}{\sqrt{D}}).
\end{equation}
This operation, however, requires constructing an \( L \times L \) attention matrix, leading to \( O(L^2) \) time and memory complexity, which becomes prohibitive for long sequences.

FlashAttention \cite{dao2022flashattention, dao2023flashattention,shah2025flashattention} addresses this issue by performing attention in a blockwise manner. Instead of storing the full attention matrix, FlashAttention processes smaller chunks sequentially. In FlashAttention, attention is computed for smaller blocks of tokens, and the key idea is to perform attention on these blocks without constructing the entire attention matrix at once. Specifically, for each block, the attention is computed as:
\begin{equation}
\mathbf{W}_{\text{attn}}^{(b)} = \text{online\_softmax}\left(\frac{(\mathbf{Q}_b \mathbf{K}_b^\top)}{\sqrt{D}}\right),
\end{equation}
where \( \mathbf{Q}_b \) and \( \mathbf{K}_b \) represent the query and key matrices for block \( b \), where $L \gg b$, and online\_softmax~\cite{milakov2018onlinenormalizercalculationsoftmax} is a blockwise equivalent version of the safe softmax. The result is then multiplied by the value matrix for the block, \( \mathbf{V}_b \), to obtain the final attention output:
\begin{equation}
\mathbf{A}_b = \mathbf{W}_{\text{attn}}^{(b)} \mathbf{V}_b.
\end{equation}
This block-wise computation  significantly reduces the memory footprint to $O(Lb)$, as only a subset of the full attention matrix is processed at any given time. 
FlashAttention is particularly effective for large-scale transformers and long-sequence tasks, such as 3D Full Attention.

\subsection{Sparse Attention and Sparse Patterns}
\label{subsec:sparse attn and sparse patterns}
Attention mechanisms exhibit inherent sparsity \cite{child2019generating}, enabling computational acceleration by limiting interactions to a subset of key-value pairs. Sparse attention reduces complexity by ignoring interactions where the attention weight \( \mathbf{W}_{\text{attn}}^{(i,j)} \) is small. This principle forms the basis of sparse attention.

Formally, sparse attention is defined by a masking function \( \mathbf{M} \in \{0,1\}^{L \times L} \), which \( \mathbf{M}_{ij} = 1 \) indicates that token \( i \) attends to token \( j \), and \( \mathbf{M}_{ij} = 0 \) removes the interaction. This masking function \( \mathbf{M} \) is \emph{sparse pattern}, the indices set of \( \mathbf{M}_{ij} = 1 \) is \emph{sparse indices}, and  the proportion of \( \mathbf{M}_{ij} = 0 \) is called \emph{sparsity}. 
The sparse attention operation is defined as:
\begin{equation}
\mathbf{A}_{\text{attn}} = \text{softmax} \left(\frac{(\mathbf{Q} \mathbf{K}^\top) \odot \mathbf{M}}{\sqrt{D}} \right) \mathbf{V},
\end{equation}
where \( \odot \) denotes element-wise multiplication. The effectiveness of a sparse pattern is evaluated using \emph{Recall}~\cite{treviso2022predictingattentionsparsitytransformers}, which measures how well the sparse pattern preserves the original dense attention behavior:
\begin{equation}
Recall = \frac{\sum_{(i,j) \in sparse \ indices} \mathbf{W}_{\text{attn}}^{(i,j)}}{\sum_{i,j} \mathbf{W}_{\text{attn}}^{(i,j)}},
\end{equation}
Higher \emph{Recall} indicates better retention of the original attention structure.







\section{Sparse Pattern Characteristic in DiTs}
\label{sec:observation}


\begin{figure*}[!t]
    \centering
    \includegraphics[width=\linewidth]{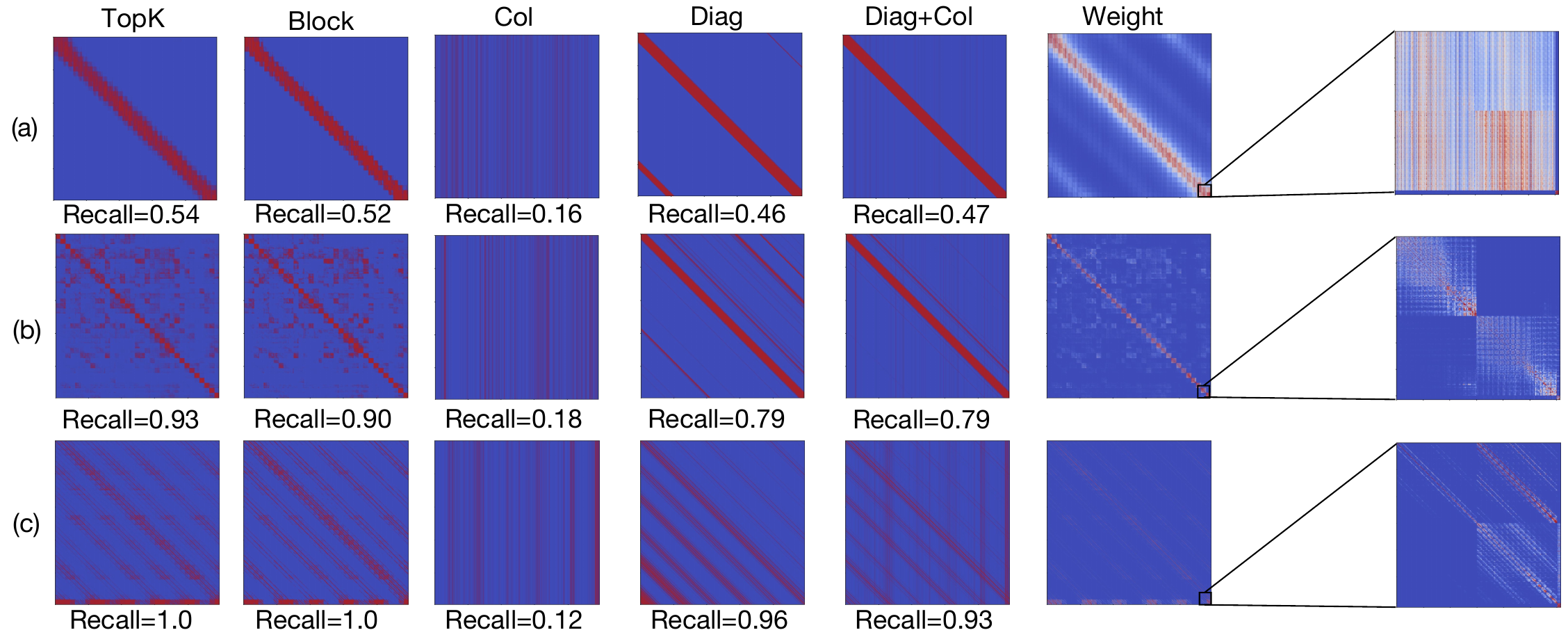}
    \caption{Typical attention weight maps from HunyuanVideo. \underline{\textit{Weight}} represents the visualization result of attention weights. \underline{\textit{Topk}}, \underline{\textit{Block}}, \underline{\textit{Col}}, \underline{\textit{Diag}}, \underline{\textit{Diag+Col}} represent the visualization results of sparse patterns under \emph{sparsity} 0.9. The far right shows an enlarged view of the attention weights selected from the bottom right corner with a size of $(2*h*w+t) \times (2*h*w+t)$, where a clear hierarchical effect between frames can be observed. At the same time, there is a distinct boundary between the text modality and the pure video modality, exhibiting varying degrees of text sink effect. (720p, 129 frames, block size of the block pattern is 32)}
    \label{fig:heatmaps}
\end{figure*}

\begin{figure*}[!t]
    \centering
    \begin{subfigure}[t]{0.28\linewidth}
        \centering
        \includegraphics[width=\linewidth]{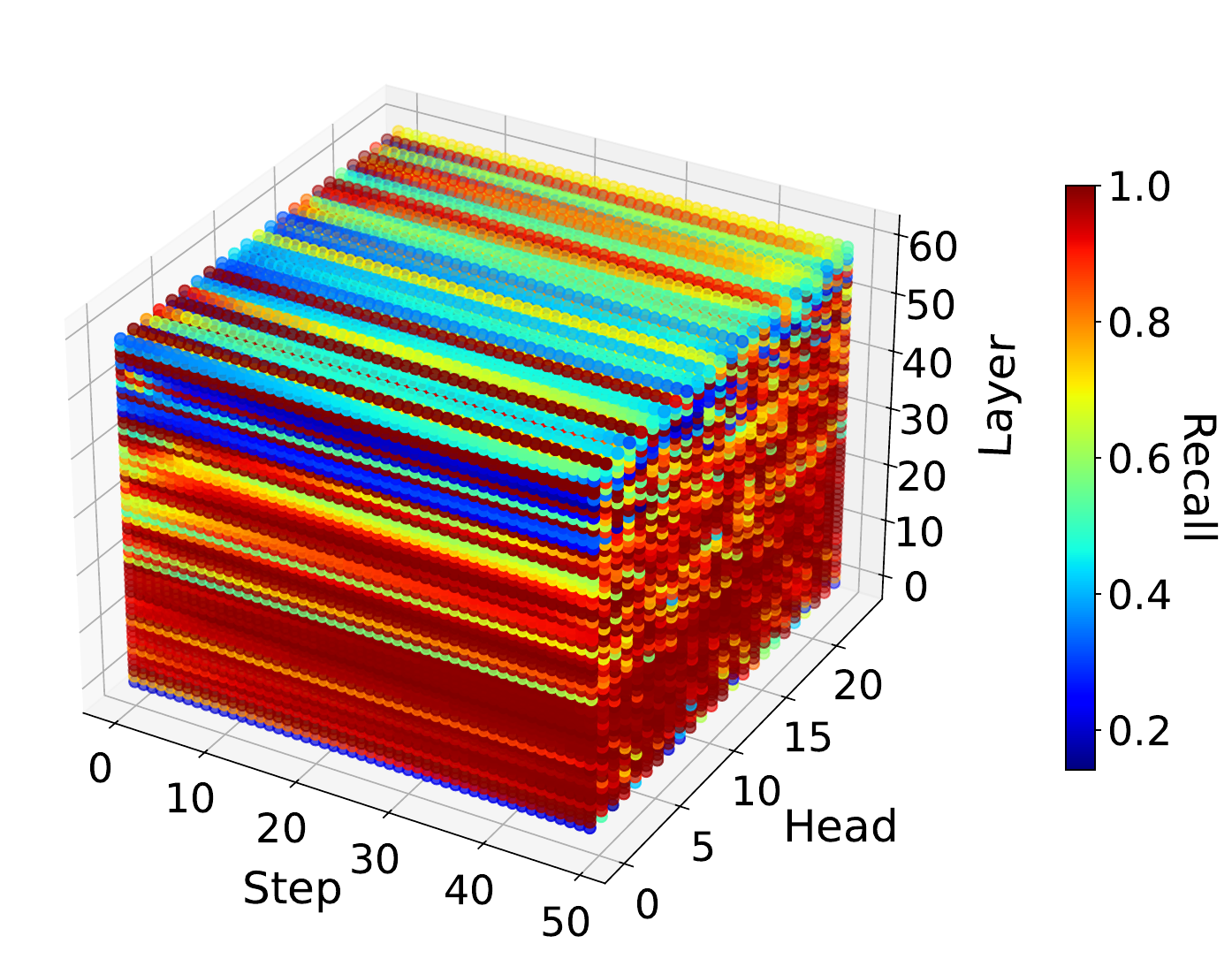}
        \caption{Sparse pattern's \emph{Recall} changes with head and layer, but invariant step.}
        \label{fig:head_layer_step_recall}
    \end{subfigure}
    \quad
    \begin{subfigure}[t]{0.28\linewidth}
        \centering
        \includegraphics[width=\linewidth]{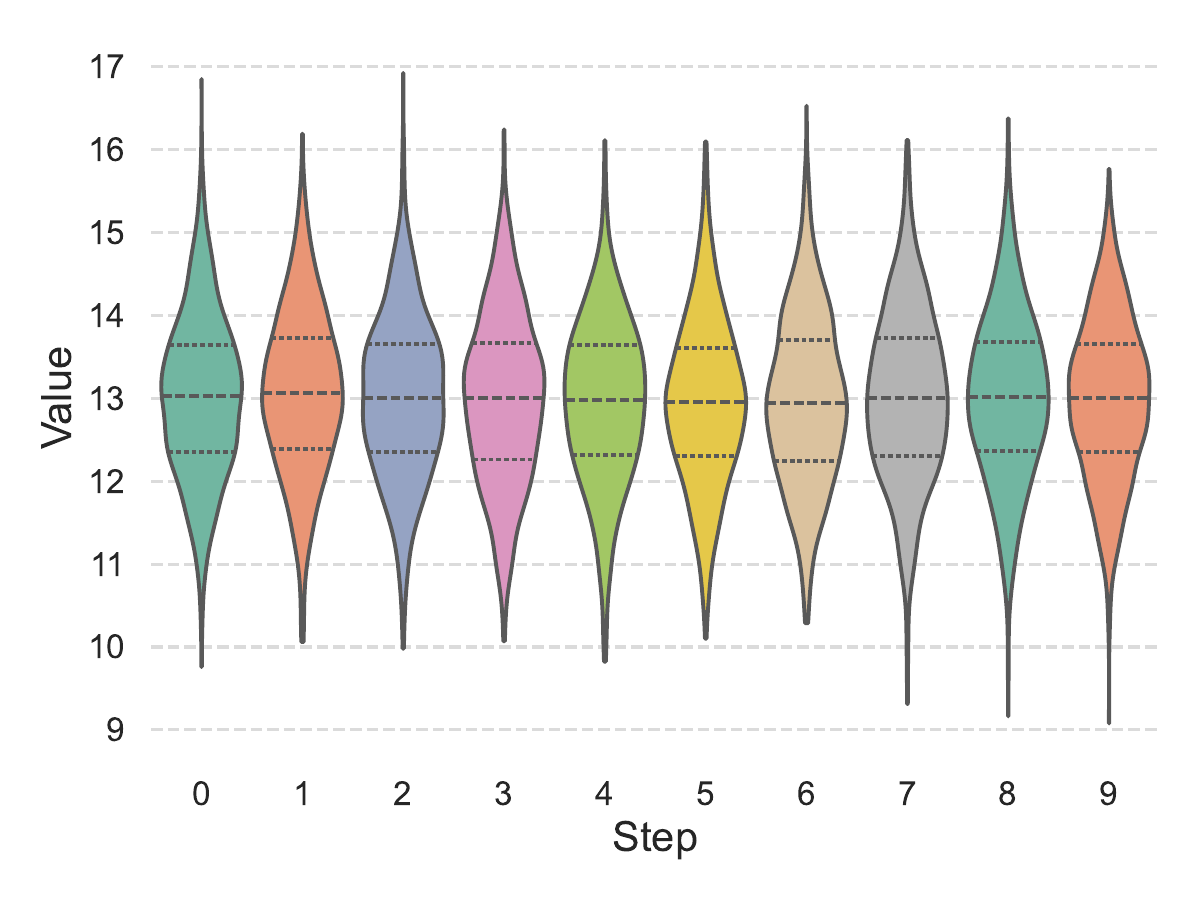}
        \caption{LSE distribution changes with the variation of step.}
        \label{fig:lse_reuse}
    \end{subfigure}
    \quad
    \begin{subfigure}[t]{0.28\linewidth}
        \centering
        \includegraphics[width=\linewidth]{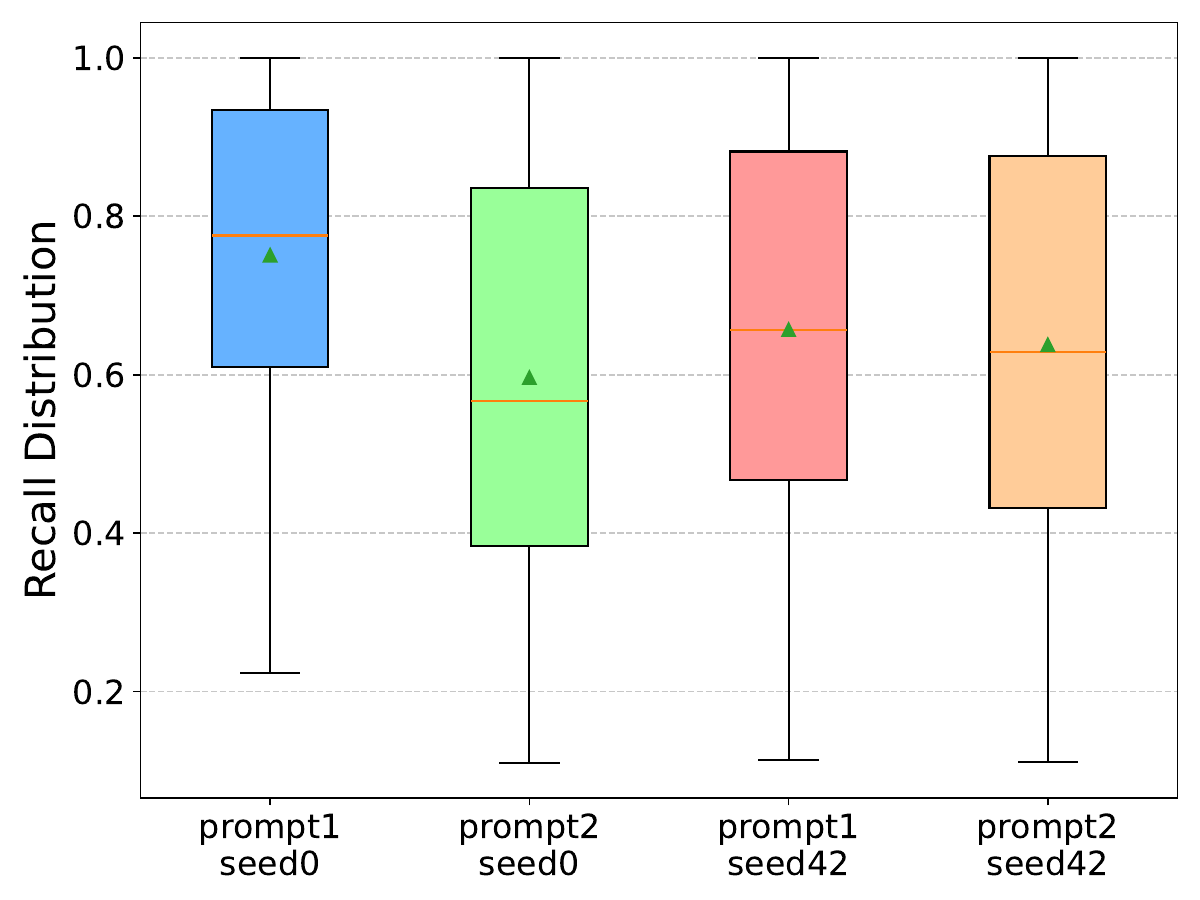}
        \caption{Sparse pattern changes with inputs
        }
        \label{fig:prompt_recall}
    \end{subfigure}
    
    \caption{(a) Visualization of recall changes with head, layer, and step. Under the condition of fixed sparsity = 0.9, the attention recall of HunyuanVideo in the \textit{TopK} paradigm changes with the variations of Head and Layer, but stay invariant with Step. (b) LSE distribution among different steps. We used HunyuanVideo to generate a 720p 8s video and recorded the distribution of LSE at each layer. It is easy to see that as the Step changes, the distribution of LSE remains almost unchanged. (c) Recall Distribution of Different Inputs. We used the best sparse pattern obtained from prompt1-seed0 and applied it to different prompts and seeds. The recall decreases when the prompt or seed changes, meaning different inputs do not share the same sparse pattern. 
    }
    \label{fig:three-subfigures}
\end{figure*}

In this section, we present the key observations of the sparse characteristics and opportunities in DiTs that motivate our work. 

\textit{\textbf{Observation 1: DiTs exhibit Hierarchical Structure of sparse pattern within and between different Modality, making continous patterns unsuitable.}}
As introduced in Section~\ref{sec:pre}, DiTs leverage 3D attention to model spatial and temporal dependencies across video frames while integrating text tokens for joint attention. 
Given an input sequence, it comprises video tokens and text tokens, with a total length of $L = f \cdot h \cdot w + t$ (Equation~\ref{eq:seqlen}).
Thus, the attention weights matrix, $\mathbf{W}_{\text{attn}} \in \mathbb{R}^{L \times L}$, has a hierarchical organization of text and video tokens. 
Particularly, as depicted in Figure~\ref{fig:heatmaps}, it can be decomposed as follows:
\begin{equation}
\mathbf{W}_{\text{attn}} = 
\begin{bmatrix}
\mathbf{W}_{\text{video-video}} & \mathbf{W}_{\text{video-text}} \\
\mathbf{W}_{\text{text-video}} & \mathbf{W}_{\text{text-text}}
\end{bmatrix},
\end{equation}
where:
\begin{itemize}
    \item \textbf{Video-video attention}, $\mathbf{W}_{\text{video-video}} \in \mathbb{R}^{(f \cdot h \cdot w) \times (f \cdot h \cdot w)}$, captures spatial and temporal interactions among video tokens.
    \item \textbf{Text-video and text-text attention}, $\mathbf{W}_{\text{text-text}} \in \mathbb{R}^{t \times t}$, $\mathbf{W}_{\text{text-video}} \in \mathbb{R}^{t \times (f \cdot h \cdot w)}$ and $\mathbf{W}_{\text{video-text}} \in \mathbb{R}^{(f \cdot h \cdot w) \times t}$, model interactions involving text tokens, which often serve as a global text sink for attention.
\end{itemize}

Moreover, within $\mathbf{W}_{\text{video-video}}$, attention weights are further structured into $f \times f$ \emph{frame regions}:
\begin{equation}
\mathbf{W}_{\text{video-video}} = 
\begin{bmatrix}
\mathbf{R}_{1,1} & \mathbf{R}_{1,2} & \cdots & \mathbf{R}_{1,f} \\
\mathbf{R}_{2,1} & \mathbf{R}_{2,2} & \cdots & \mathbf{R}_{2,f} \\
\vdots & \vdots & \ddots & \vdots \\
\mathbf{R}_{f,1} & \mathbf{R}_{f,2} & \cdots & \mathbf{R}_{f,f}
\end{bmatrix},
\end{equation}
where $\mathbf{R}_{i,j} \in \mathbb{R}^{(h \times w) \times (h \times w)}$ represents interactions between the $i$-th and $j$-th video frames. 
As shown in Figure~\ref{fig:heatmaps}, there are clear boundaries between the frames.

This hierarchical characteristic makes continuous sparse patterns ineffective, as the sparsity structure is no longer globally uninterruptible. 
In a continuous sparse pattern, nonzero elements extend continuously across the entire matrix, such as \emph{col} patterns, where specific columns remain active in all rows, or \emph{diag} patterns, where nonzero values form a diagonal path from one side to the other. 
However, due to the hierarchical structure of certain attention weight, their sparse patterns become fragmented rather than maintaining such continuity, making it impossible to describe them using continuous sparse patterns. 
Nevertheless, while the overall structure lacks continuity, we observe that within each frame region, the sparsity pattern remains locally structured and can often be well characterized using continuous patterns like \emph{col} or \emph{diag}. 

This insight motivates a \emph{frame region}-wise search strategy to capture localized continuous structures and reconstruct the overall sparsity pattern. However, as shown in Figure~\ref{fig:heatmaps}, attention distribution varies significantly across different \emph{frame regions}, nonzero weights tend to concentrate in only a few \emph{frame regions} rather than being evenly distributed. 
This imbalance reduces the effectiveness of the frame region-wise approach, as it fails to provide a globally optimized sparse representation.

\textit{\textbf{Solution 1: Using the blockified pattern to describe the sparse features of DiT.}}
Although continuous patterns like col or diag do not work well, we find that the \emph{sparse pattern} evolves into a \textit{blockified} structure globally. For example, as shown in Figure~\ref{fig:heatmaps}a, within each \emph{frame region}, the sparsity follows a \textit{col} pattern. However, due to weak inter-region interactions, hierarchical sparsity disrupts interlinearly continuous \emph{col} patterns, leading to a \textit{blockified} structure. As observed in the figure, this \textit{blockified} characteristic achieves better \emph{Recall}, indicating the \emph{blockified} pattern a more suitable pattern. Similarly, in Figure~\ref{fig:heatmaps}b, each \emph{frame region} follows a hybrid of \textit{diag} and \textit{col} patterns. Yet, due to significant variations in inter-frame interactions, the global attention weights exhibit a combination of a \emph{sliding window} pattern and a distinct \textit{random blockified} structure, making it impossible to describe with standard sparsity patterns. Another example is shown in Figure~\ref{fig:heatmaps}c, where individual \emph{frame regions} lack a clear local pattern, while the global attention weights form a \emph{noncontinuous-diag} pattern combined with a \emph{bottom sink} effect. As seen in Figure~\ref{fig:heatmaps}c, this characteristic can also be effectively modeled using a \emph{blockified} representation with the best \emph{Recall}.

In summary, due to the hierarchical nature of the DiT patterns, conventional continuous patterns fail to provide an effective representation. Thus, adopting the \emph{blockified} pattern is the optimal choice for capturing the sparsity characteristics of DiT, because it consistently achieves the best recall, as shown in Figure~\ref{fig:heatmaps}.  





\textit{\textbf{Observation 2: DiTs' sparse pattern vary w.r.t. inputs, layers and heads, making offline search unsuitable.}}
As illustrated in Figure~\ref{fig:head_layer_step_recall}, the sparse patterns in DiTs vary depending on attention head, and layer, which is similar to LLMs~\cite{jiang2024minference,liu2021transformer}.

Meanwhile, we observe that the sparse patterns of different prompts also vary significantly. In Figure~\ref{fig:prompt_recall}, we conducted the following experiment: we searched for the optimal sparse pattern for a specific prompt with a fixed \emph{sparsity} of 0.9. Subsequently, this pattern was directly applied to other prompts. We selected various prompts and different random seeds, and the results revealed that the sparse pattern optimized for one input is not necessarily optimal for other inputs.
These observations reveal that the sparse patterns of different prompts differ significantly, making offline searches likely to fail. 

Another conventional approach is \emph{online approximate search}~\cite{jiang2024minference}. However, due to the hierarchical structure and dispersed attention distribution described in Observation 1, this method fails to accurately capture the correct sparse indices, resulting in poor performance within DiT (as evaluated in Section~\ref{sec:experiments}). 

Therefore, DiT requires a  \emph{precise online search}; however, its prohibitive computational cost makes it impractical, which is why no prior methods have adopted it.  

\textit{\textbf{Solution 2: DiTs' sparse pattern and LSE keep invariant in diffusion steps, caching those invariables making a fast precise online search feasible.}}
DiTs perform an iterative multi-step denoising process, and we observe an important invariance: for a given layer and head, although the specific values of the attention weights change dynamically across denoising steps, the underlying sparse pattern remains consistent throughout the process. 
Furthermore, we statistically analyze the distribution of the LSE data calculated in FlashAttention at different steps within the same layer. 
The results in Figure~\ref{fig:lse_reuse} show that the distribution of LSE remains stable across denoising steps.

Those similarities between consecutive steps provide an opportunity to explore the reuse of sparse patterns and LSE to accelerate online search, as detailed in Section~\ref{sec:method}.

\section{Methodology}
\label{sec:method}




Motivated by those observations, we propose \textbf{AdaSpa}, a sparse attention mechanism featuring Dynamic Pattern and Online Precise Search, to accelerate long video generation with DiTs. 

\begin{figure*}[!t]
    \centering
    \includegraphics[width=\linewidth]{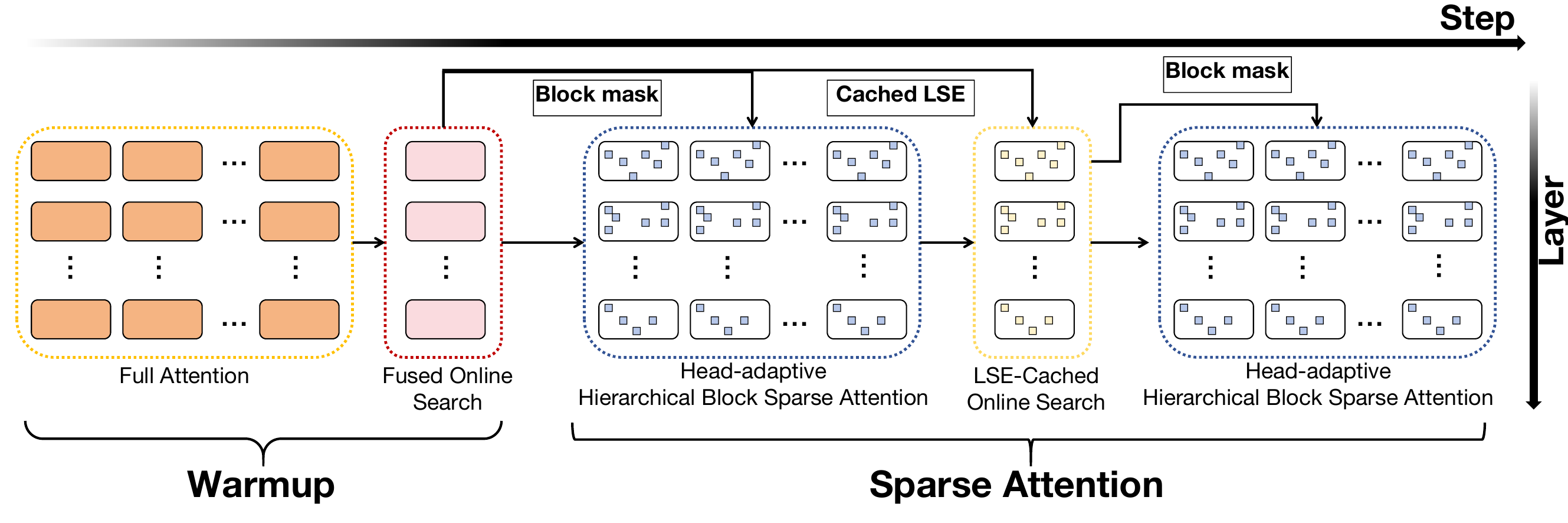}
    \caption{Overview of AdaSpa. 
    We define a warm-up step $T_w$ = $\{1, 2, ..., t_w\}$, and select k steps: $T_{s} = \{t_{s}^1, t_{s}^2, ..., t_{s}^k \}$ to perform a precise online search, with $t_{key}^1 = t_w$. Initially, during steps $1$ to $t_w - 1$, we use full attention. At step $t_w$, we apply Fused Online Search to do full attention and thereby compute block mask, which is then passed to the subsequent steps $t_{key}^1+1, t_{key}^1+2, \dots, t_{key}^2-1$ for Head-adaptive Hierarchical Block Sparse Attention.
    Subsequently, for each $t_{key}^i$, where $i > 1$, we leverage the Cached LSE from the previous $t_{key}^1$ search to perform the LSE-Cached Online Search, thereby obtaining a new mask. This new mask is then passed to the subsequent steps $t_{key}^i, t_{key}^i+1, t_{key}^i+2, \dots, t_{key}^{i+1}-1$ for Head-adaptive Hierarchical Block Sparse Attention computation.
    }
    \label{fig:overview}
\end{figure*}


\subsection{Problem Formulation}

Section~\ref{sec:observation} demonstrates that the attention weights of DiTs cannot be well represented using patterns such as \emph{col} or \emph{diag} due to the discontinuities caused by hierarchical structures, while the \emph{block} pattern shows advantages.
Thus, to facilitate the online search of dynamic sparse masks, we formulate the problem of how to find the optimal block sparse indices.

\textbf{Definition of Blockified Sparse Attention.} Block Sparse Attention employs a block-wise attention method similar to FlashAttention, with the distinction that Block Sparse Attention ignores the computation of certain blocks based on its sparse indices, thereby achieving a speedup.
Concretely, partition the length dimension $L$ into $L/B$ chunks, where $B$ is the block size of sparse attention, and define a block-level sparse pattern $\mathbf{M}_{S}\in\{0,1\}^{\tfrac{L}{B}\times\tfrac{L}{B}}$, 
where $S$ is the set of sparse indices of $\mathbf{M}$.
By expanding $\mathbf{M}_S$ to $\widetilde{\mathbf{M}_S}\in\{0,1\}^{L\times L}$ and applying a large negative bias $-\,c\,(1-\widetilde{\mathbf{M}_S})$, we can exclude the discarded blocks from the safe Softmax computation:
\begin{equation}
\mathbf{W}_{\text{attn}}(\widetilde{\mathbf{M}_S})
\;=\;
\mathrm{Softmax}_{\mathrm{safe}}\!\Bigl(
\tfrac{\mathbf{Q}\mathbf{K}^\top}{\sqrt{D}}
\;-\;c\bigl(1-\widetilde{\mathbf{M}_S}\bigr)\Bigr),
\end{equation}
where $c$ is sufficiently large. 

\textbf{Optimal sparse indices.} The goal of block sparse attention is to retain as much of the attention weights as possible, thus to achieve the best \emph{Recall}.

We predefine $\mathbf{W}_{\text{sum\_attn}}$ as the sum of attention weights within each block of $\mathbf{W}_{\text{attn}}$:
\begin{equation}
\mathbf{W}_{\text{sum\_attn}} = \sum_{i=0}^{B-1} \sum_{j=0}^{B-1} \mathbf{W}_{\text{attn}}[B \cdot p + i, B \cdot q + j]
\end{equation}
where $p, q \in \{0, 1, \dots, \frac{L}{B}-1\}, \mathbf{W}_{\text{sum\_attn}} \in \mathbb{R}^{\frac{L}{B} \times \frac{L}{B}}$

Formally, at a given \emph{sparsity}, the precise \emph{sparse indices} of block sparse attention can be expressed as: 

\begin{equation}
\begin{aligned}
S^* &= \arg\min_{S} \;\bigl\|\mathbf{W}_{\text{attn}} - \mathbf{W}_{attn}(\widetilde{\mathbf{M}_S})\bigr\| \\
&= \arg\max_{S} \;\bigl\|\mathbf{W}_{attn}(\widetilde{\mathbf{M}_S})\bigr\| \\
&= \arg\max_{S} \;\bigl\|\mathbf{W}_{sum\_attn}(\mathbf{M}_S)\bigr\| \\
&= \arg\max_{k \in \{1, \dots, (1- \text{sparsity})(\frac{L}{B})^2\}} \mathbf{W}_{sum\_attn}[k]
\end{aligned}
\end{equation}
This indicates that we can obtain the optimal \emph{sparse indices} by calculating $\mathbf{W}_{sum\_attn}$ and utilizing topk.
We only need to calculate the block with index in $S^*$, while omitting other blocks. 
Thus, under the given sparsity, the complexity can be reduced from $\mathcal{O}(L^2 d)$ to $\mathcal{O}((1-sparsity) L^2 d)$, providing a significant speedup.


\subsection{Design of Adaptive Sparse Attention}

We illustrate the overview of AdaSpa in Figure~\ref{fig:overview}. As previously mentioned, in order to perform a precise search, it is necessary to obtain the complete $\mathbf{W}_{\text{attn}}$, which has a size of $O(L^2)$. In the context of long video generation, this results in significant time and memory overhead. Moreover, since the mask for each attention operation must be determined in real-time, this time consumption is not affordable~\cite{jiang2024minference}.

To address this issue, we exploit the property of DiT's sparse pattern, which exhibits similarity in denoising steps, and construct AdaSpa with a two-phase Fused LSE-Cached Online Search and Head-adaptive Hierarchical Block Sparse Attention.

\begin{algorithm}[!t]
\caption{Fused Online Search}
\label{alg:fused_online_search}
\begin{algorithmic}[1]

\INPUT 
$Q, K, V$,
\OUTPUT
$LSE, Out, W_{\text{sum\_attn}}$

\State Initialize: $lse \gets -\infty$, $row\_max \gets 1$, $acc \gets 0$
\State Load query block in parallel: $q \gets Q[\text{current block}]$
\State \textit{// First Pass: Compute FlashAttention outputs and store LSE.}

\For{\textbf{each} key block $k \in K$, value block $v \in V$}
    \State $qk \gets \text{Dot}(q, k)$
    \State $row\_max \gets \text{update}(row\_max, qk)$
    \State $p \gets \text{online\_softmax}(row\_max, qk)$
    \State $lse += \text{Sum}(p, -1)$
    
    \State $acc \gets \text{Dot}(p, v, acc)$
\EndFor
\State $LSE \gets \text{Log}(lse)+row\_max$
\State $Out \gets acc$

\vspace{0.5em}

\State \textit{// Second Pass: Use cached LSE to compute $W_{sum\_attn}$ and reduce time.}
\For{\textbf{each} key block $k \in K$}
    \State $qk \gets \text{Dot}(q, k)$
    \State $p \gets \text{Log}(qk -LSE)$
    \State $p\_sum = \text{Sum(p)}$
    \State Store $p\_sum$ to coresponding position in $W_{\text{sum\_attn}}$ 
\EndFor

\State \textbf{Return:} $LSE$, $Out$, $W_{\text{sum\_attn}}$

\end{algorithmic}
\end{algorithm}

\begin{algorithm}[!t]
\caption{LSE-Cached Online Search}
\label{alg:lse_cached_online_search}
\begin{algorithmic}[1]

\INPUT 
$Q, K, LSE$,
\OUTPUT
$W_{\text{sum\_attn}}$
\State Load query block in parallel: $q \gets Q[\text{current block}]$
\State \textit{// Only one pass; use cached LSE to compute $W_{sum\_attn}$.}
\For{\textbf{each} key block $k \in K$}
    \State $qk \gets \text{Dot}(q, k)$
    \State $p \gets \text{Log}(qk -LSE)$
    \State $p\_sum = \text{Sum(p)}$
    \State Store $p\_sum$ to coresponding position in $W_{\text{sum\_attn}}$ 
\EndFor

\State \textbf{Return:} $W_{\text{sum\_attn}}$

\end{algorithmic}
\end{algorithm}

\textbf{Fused LSE-Cached Online Search.} The first phase of Fused LSE-Cached Online Search is a Fused online Search, which is a two-pass search: the first pass computes the original FlashAttention outputs and stores each row’s LSE, while the second pass uses the previously generated LSE to
compute $\mathbf{W}_{sum\_attn}$ in a block-wise manner fused with FlashAttention.

The second phase is an LSE-Cached online Search, which only contains one pass. Due to the similarity of LSE in steps, we directly use the LSE obtained from the Fused online Search to calculate $\mathbf{W}_{sum\_attn}$, thereby saving one pass of search time and further reducing the search time by half. Algorithm~\ref{alg:fused_online_search} and~\ref{alg:lse_cached_online_search} demonstrate the pseudocode of our precise online search.

\textbf{Head-adaptive Hierarchical Block Sparse Attention.} Figure~\ref{fig:head_layer_step_recall} shows that not all attention heads share the same sparsity characteristics. 
A single uniform sparsity across all heads is often suboptimal because certain heads may function well with fewer retained blocks, while others require more. However, if each head employs a totally distinct sparsity level, it will cause huge search time and lead to severe kernel load imbalance that significant wastage of computational resources.

To utilize the head adaptive feature while mitigate wastage of computational resources, we employ a hierarchical search and calculation strategy. 
Specifically, we start by fixing a given \emph{sparsity} and computing the \emph{Recall} for each head. We then sort all heads according to their respective \emph{Recall}. Let $n$ denote the number of heads whose \emph{Recall} exceeds $0.8$, which is a well-known fine \emph{Recall} to a sparse attention. 
Next, we increase the \emph{sparsity} of the $n$ heads with the highest \emph{Recall} to $\frac{1 + \text{sparsity}}{2}$, and we decrease the \emph{sparsity} of the $n$ heads with the lowest \emph{Recall} to $\frac{3 \times \text{sparsity} - 1}{2}$. 
This hierarchical head-adaptive procedure effectively reduces redundancy among heads exhibiting higher \emph{Recall} while improving the precision of those with lower \emph{Recall}. Consequently, we achieve elevated accuracy without altering the average sparsity, thus realizing a head-adaptive mechanism. 

\subsection{Implementation}
\begin{figure}[!t]
    \centering
    \includegraphics[width=\linewidth]{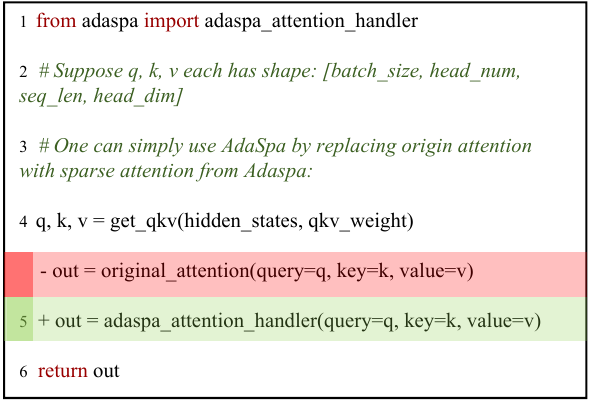}
    \caption{Minimal usage of \texttt{AdaSpa}.}
    \label{list:adaspa_example}
\end{figure}

AdaSpa is implemented with over 2,000 lines of Python and 1000 lines of Triton~\cite{10.1145/3315508.3329973} codes. It is provided as a plug-and-play interface, as shown in Figure~\ref{list:adaspa_example}.
Users can enable AdaSpa with only a one-line change. 
We use sparsity=0.8, block\_size=64, $T_s = \{10, 30\}$ as the default configuration. We implement our Head-adaptive Hierarchical Block Sparse Attention based on Block-Sparse-Attention~\cite{guo2024blocksparse}. Unless otherwise noted, all other attention mechanisms employ FlashAttention 2~\cite{dao2023flashattention}.

In addition, we employ two optimization techniques for better efficiency. (1) \emph{Text Sink.} We manually select all the indices of video-text, text-video, and text-text parts, which can enhance video modality's perception to text modality, thereby achieving better results.
(2) \emph{Row Wise.} We find that ensuring each query attends to roughly the same number of keys can improve continuity in generated videos. Otherwise, certain regions deemed ``unimportant'' might never be attended to, producing artifacts. Hence, we enforce a \emph{per-row uniform selection} in our block \emph{sparse pattern}.





\section{Experiments}
\label{sec:experiments}
\noindent
\textbf{Models.} 
We experiment with two state-of-the-art open-source models, namely \textit{HunyuanVideo} (13B)~\cite{kong2024hunyuanvideo} and \textit{CogVideoX1.5-5B}~\cite{yang2024cogvideox}. We generate 720p, 8-second videos for HunyuanVideo, 720p and 10-second videos for CogVideoX1.5-5B, with 50 steps for both of these models. 

\noindent
\textbf{Baselines.}
We compare AdaSpa with \textit{Sparse VideoGen}~\cite{xi2025sparse} (static pattern) and \textit{Minference}~\cite{jiang2024minference} (dynamic pattern). 
In addition, we also consider two variants of AdaSpa to assess the effectiveness of the proposed methods: 
(1) \textit{AdaSpa (w/o head adaptive)}, with uses the same sparsity for all heads, and (2) \textit{AdaSpa (w/o lse cache)}, which does not leverage the LSE-Cached method. 
For all methods, the first 10 steps generate with full attention for warmup.

\noindent
\textbf{Datasets.}
For all the experiments, we use the default dataset from VBench~\cite{huang2023vbench} for testing. Specially, for \textit{CogVideoX1.5-5B}, we use VBench dataset after applying prompt optimization, following the guidelines provided by CogVideoX~\cite{yang2024cogvideox}.

\noindent
\textbf{Metrics.}
To evaluate the performance of our video generation model, we employ several widely recognized metrics that assess both the quality and perceptual similarity of the generated videos. Following previous works~\cite{li2024svdqunat,zhao2024real,kahatapitiya2024adaptive}, we utilize Peak Signal-to-Noise Ratio~\cite{opencv_github} (PSNR), Learned Perceptual Image Patch Similarity~\cite{zhang2018unreasonableeffectivenessdeepfeatures} (LPIPS), and Structural Similarity Index Measure~\cite{1284395} (SSIM) to evaluate the similarity of generated videos.
As for video quality, we introduce the VBench Score~\cite{huang2023vbench}, which provides a more comprehensive evaluation by considering both pixel-level accuracy and perceptual consistency across frames. 
For efficiency, we report latency and speedup, where both are measured using a single A100 GPU-80GB.

\begin{table*}[t!]
\centering
\begin{tabular}{|l|cccc|cc|}

\hline

\multicolumn{1}{|c|}{\multirow{2}{*}{\textbf{Method}}} & \multicolumn{4}{c|}{\textbf{Quality Metrics}} & \multicolumn{2}{c|}{\textbf{Efficiency Metrics}} \\

\cline{2-7}

& \textbf{VBench (\%)} $\uparrow$ & \textbf{PSNR} $\uparrow$ & \textbf{SSIM} $\uparrow$ & \textbf{LPIPS} $\downarrow$ & \textbf{Latency (s)} & \textbf{Speedup} \\

\hline

\rowcolor{gray!10}
HunyuanVideo & 80.10 & - & - & - & 3213.76 & 1.00× \\
+ MInference  & 79.17 & 22.53 & 0.7435 & 0.3550 & 2532.80 & 1.27× \\
+ Sparse VideoGen  & 79.39 & 27.61 & 0.8683 & 0.1703 & 2035.59 & 1.58× \\
\rowcolor[HTML]{E4F4FC}
+ AdaSpa (w/o head adaptive) & 79.64 & 28.51 & 0.8825 & 0.1574 & 1823.34 & 1.76×  \\
\rowcolor[HTML]{E4F4FC}
+ AdaSpa (w/o lse cache) & \textbf{80.16} & 28.97 & 0.8898 & 0.1481 & 1877.13 & 1.71×  \\
\rowcolor[HTML]{E4F4FC}
+ AdaSpa (\textbf{ours}) & 80.13 & \textbf{29.07} & \textbf{0.8905} & \textbf{0.1478} & \textbf{1810.23} & \textbf{1.78×} \\

\hline

\rowcolor{gray!10}
CogVideoX1.5 & 81.16 & - & - & - & 3135.24 & 1.00× \\
+ MInference  & 65.30 & 10.31 & 0.3113 & 0.6820 & 2258.35 & 1.39× \\
+ Sparse VideoGen  & 79.40 & 18.98 & 0.6465 & 0.3632 & 2061.42 & 1.52× \\
\rowcolor[HTML]{E4F4FC}
+ AdaSpa (w/o head adaptive) & 81.54 & 22.99 & 0.8133 & 0.2203 & 1915.88 & 1.64× \\
\rowcolor[HTML]{E4F4FC}
+ AdaSpa (w/o lse cache) & 81.73 & 23.14 & 0.8255 & 0.2091 & 1961.71 & 1.60×  \\
\rowcolor[HTML]{E4F4FC}
+ AdaSpa (\textbf{ours}) & \textbf{81.90} & \textbf{23.25} & \textbf{0.8267} & \textbf{0.2067} & \textbf{1888.14} & \textbf{1.66×} \\

\hline

\end{tabular}
\caption{Quantitative evaluation of quality and latency for AdaSpa and other methods.}
\label{tab:main_results}
\end{table*}

\subsection{Main Results}


In Table~\ref{tab:main_results}, we present a comprehensive evaluation of AdaSpa, comparing it with various baseline methods across both quality and efficiency metrics. 

We observe that AdaSpa consistently achieves the best performance in both quality and efficiency across all experiments. On HunyuanVideo, AdaSpa ranks first in most metrics and achieves the highest speedup of 1.78$\times$. In contrast, both Sparse VideoGen and MInference show suboptimal results, with speedups of 1.58$\times$ and 1.27$\times$, respectively. On CogVideoX1.5-5B, AdaSpa delivers the best performance across all quality metrics and achieves a speedup of 1.66$\times$, the highest among the evaluated methods.

MInference, due to its reliance on online approximate search, struggles to accurately capture the precise sparse indices, leading to the lowest accuracy. 
Moreover, because of the dispersed characteristic of sparse patterns in DiT, the patterns obtained through approximate search exhibit a lower true sparsity, resulting in slower performance with speedups of only 1.27$\times$ and 1.39$\times$. 
Sparse VideoGen, which leverages a static pattern that is specifically designed for DiT, performs relatively well, as it can capture some optimal sparse patterns for specific heads. However, due to its inability to dynamically capture accurate sparse patterns for all heads, it fails to outperform AdaSpa in all accuracy metrics.

For the two variants of AdaSpa, AdaSpa (w/o head adaptive) shows worse performance in terms of quality metrics, providing strong evidence of the effectiveness of head-adaptive sparsity. Additionally, AdaSpa (w/o LSE cache) generally performs worse or on par with AdaSpa across most metrics. Due to slower search speeds, it only achieves speedups of 1.71$\times$ and 1.60$\times$ on Hunyuan and CogVideoX1.5-5B, respectively, both lower than AdaSpa’s performance. 
This further corroborates the effectiveness of LSE-Cached Search and our Head-adaptive Hierarchical method in enhancing speedup and quality.



\subsection{Ablation Study}
\begin{figure*}[!t]
    \centering
    \includegraphics[width=0.95\linewidth]{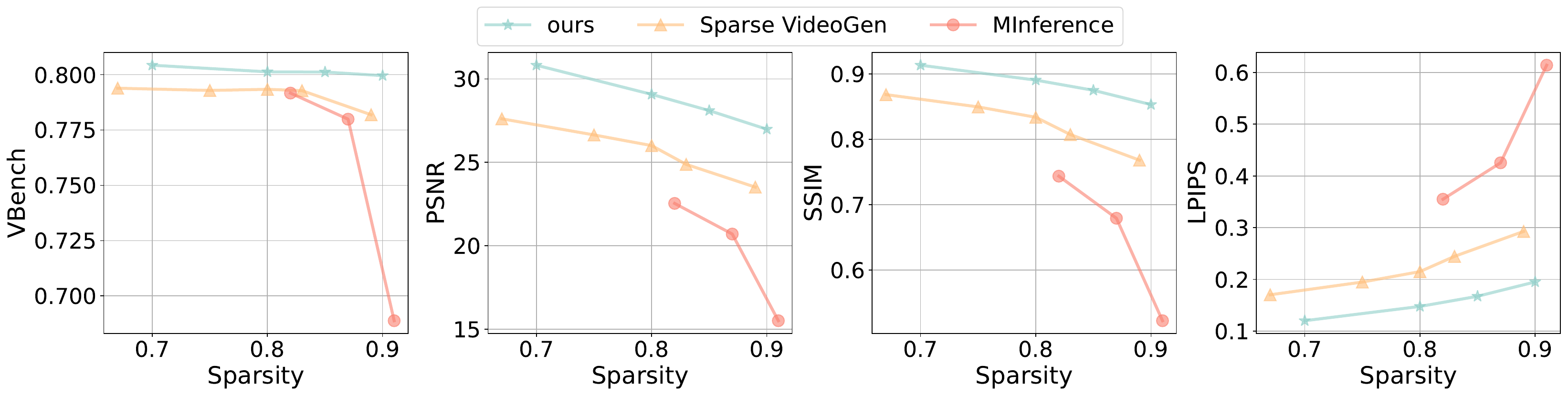}
    \caption{Quality-Sparsity trade off.
    }
    \label{fig:ablation}
\end{figure*}
\textbf{Quality-Sparsity trade-off.} In Figure~\ref{fig:ablation}, we compare the quality metrics of AdaSpa with MInference and Sparse VideoGen at different sparsity levels. 
As observed in the VBench metric, which measures video quality, AdaSpa consistently maintains the highest video quality across all sparsity levels, with no significant degradation as sparsity increases. 
In contrast, both Sparse VideoGen and MInference experience a considerable drop in quality as sparsity increases. This demonstrates that AdaSpa is capable of preserving critical information as much as possible under limited sparsity, thereby ensuring the reliability of video quality.

Similarly, in the PSNR, SSIM, and LPIPS metrics, which measure the similarity between the videos generated with and without sparse attention, a consistent trend is observed: as sparsity increases, the similarity for all video methods declines. However, AdaSpa maintains significantly higher similarity compared to other methods, with a gradual linear decrease as sparsity increases. This is in stark contrast to the abrupt decline observed in MInference.

\begin{figure}[!t]
    \centering
    \includegraphics[width=\linewidth]{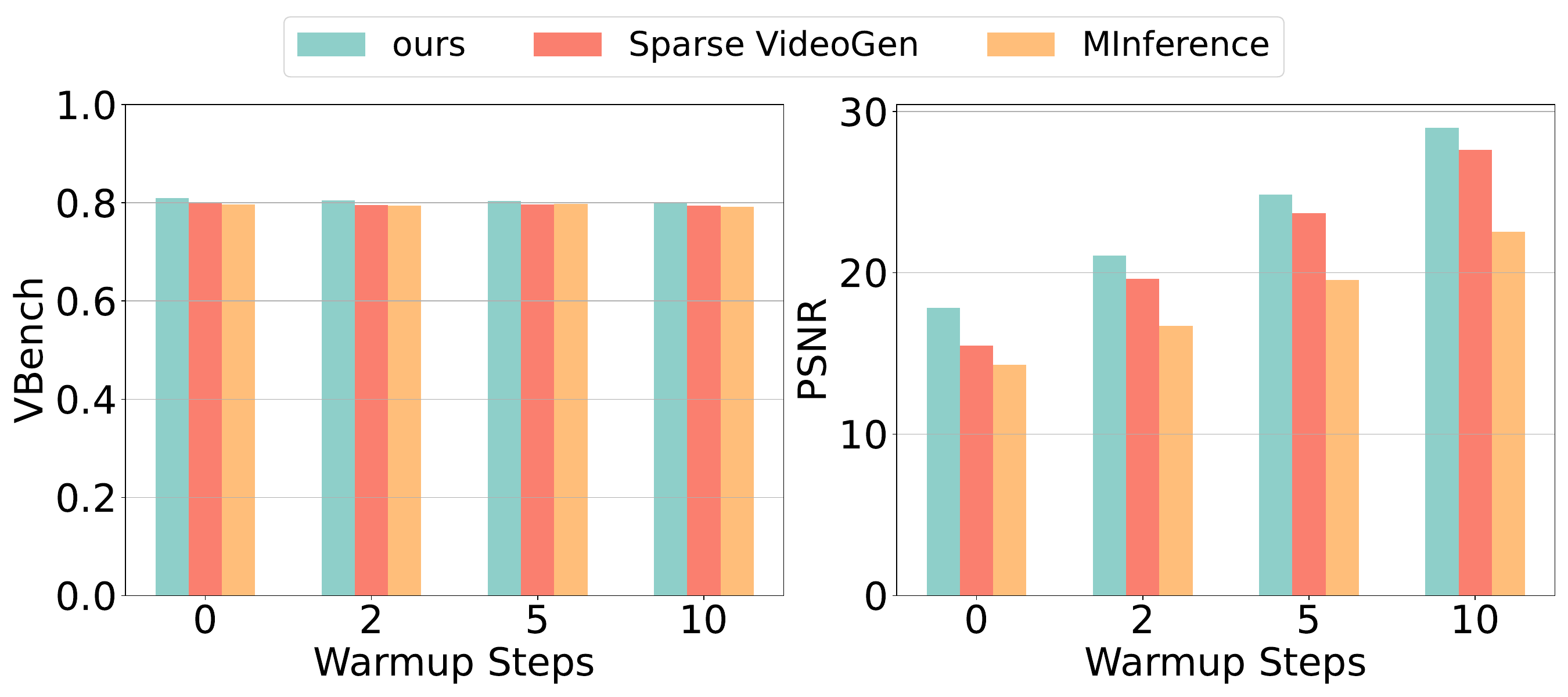}
    \caption{The impact of different warmup steps for AdaSpa, Sparse VideoGen, and MInference.}
    \label{fig:warmup}
\end{figure}

\textbf{Warmup.} As mentioned in many previous works~\cite{ma2024deepcache,kahatapitiya2024adaptive,xi2025sparse}, warmup can significantly enhance the similarity and stability of video generation. Therefore, we compared the video quality and similarity of AdaSpa, MInference, and Sparse VideoGen under different warmup setups in Figure~\ref{fig:warmup}. It can be seen that as warmup decreases, the similarity of all methods also decreases, which is consistent with the conclusions of previous works. However, we find that as the warmup period increases, AdaSpa still achieves the best performance across all setups. Additionally, the video quality for all methods does not show significant improvement with the increase in warmup, remaining almost unchanged. This suggests that warmup has minimal impact on the quality of video generation itself and primarily affects the similarity between the generated video and the original video.

\textbf{Search Strategy.}
To verify the impact of our search strategy on video generation, we evaluate AdaSpa on video quality and similarity with different search strategies, as shown in Table~\ref{tbl:search}. 
The results indicate that increasing the number of searches might be beneficial for improving accuracy, yet to a limited extent. When the number of searches reaches a certain threshold, further increasing the number of searches may even lower the video generation quality. 
This sufficiently demonstrates that the patterns between steps have a strong similarity, and as the number of searches increases, the video quality may actually decline. This suggests that the impact of sparse attention has a certain transmissibility and may affect subsequent steps, which will be further explored in future work.


\begin{table}[!t]
\centering
\small
\caption{The impact of different Search Strategies for AdaSpa.}
\begin{tabular}{c|c|c|c}
\hline
\textbf{Search Strategy ($T_s$)} & \textbf{PSNR $\uparrow$} & \textbf{SSIM $\uparrow$} & \textbf{LPIPS $\downarrow$} \\ \hline
\{10\} & 28.9629 & 0.8879 & 0.1509 \\ \hline
\{10, 30\} & 29.0749 & 0.8905 & 0.1478 \\ \hline
\{10, 20, 30\} & 28.9343 & 0.8894 & 0.1500 \\ \hline
\{10, 20, 30, 40\} & 28.9313 & 0.8898 & 0.1494 \\ \hline
\end{tabular}
\label{tbl:search}
\end{table}

\subsection{Scaling Study}

\begin{figure}[!t]
    \centering
    \includegraphics[width=0.8\linewidth]{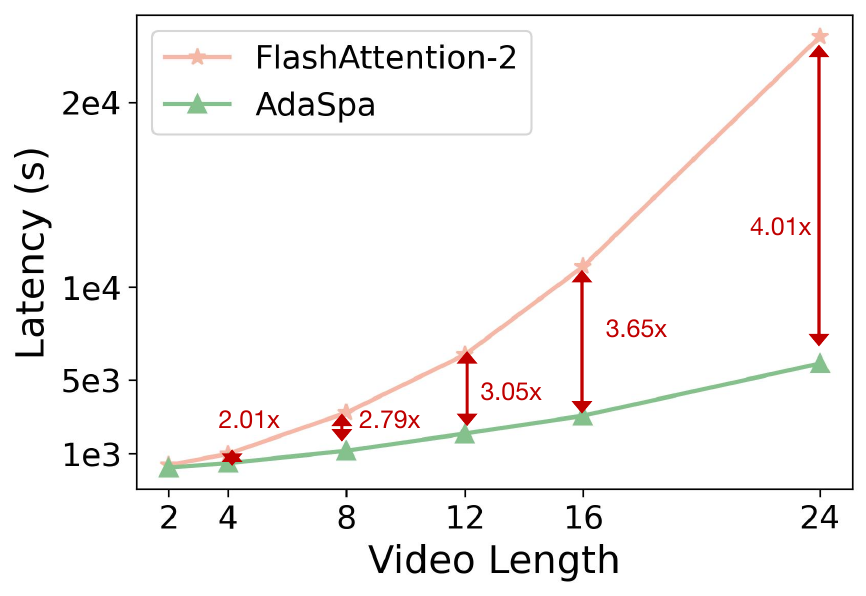}
    \caption{Scaling test of AdaSpa.}
    \label{fig:scaling}
\end{figure}

To further assess the scalability of our method, we tested the generation time for videos of different lengths under the configuration of \emph{sparsity}=0.9, block\_size=64, and $T_s=\{0, 30\}$. As shown in Figure~\ref{fig:scaling}, as the length of the generated video increases, AdaSpa’s speedup continues to improve, ultimately reaching a speedup of 4.01$\times$ when the video length is 24 seconds. This demonstrates the excellent scalability of our method. 


\section{Conclusion}
\label{sec:conc}

In this work, we comprehensively analyze the sparse characteristics in the attention mechanisms when generating videos with DiTs. 
Based on the observations and analyses, we develop AdaSpa, a brand new sparse attention approach featuring dynamic pattern and online precise search, to accelerate long video generation. 
Empirical results show that AdaSpa achieves a 1.78$\times$ of efficiency improvement while maintaining high quality in the generated videos. 

{
    \small
    \bibliographystyle{ieeenat_fullname}
    \bibliography{main}
}

\end{document}